\PassOptionsToPackage{utf8}{inputenc}
\documentclass{bioinfo}
\copyrightyear{2021} \pubyear{2021}

\usepackage{multirow}
\usepackage{graphicx}
\usepackage{paralist}
\usepackage{array}
\usepackage{supertabular,booktabs}
\usepackage{hyperref}

\access{Advance Access Publication Date: Day Month Year}
\appnotes{Original Paper}

\usepackage{adjustbox}
\usepackage{url}
\usepackage{lipsum}

\usepackage{tikz}
\def\checkmark{\tikz\fill[scale=0.4](0,.35) -- (.25,0) -- (1,.7) -- (.25,.15) -- cycle;} 

\begin{document}
\newcommand{\eat}[1]{\ignorespaces}
\firstpage{1}

\subtitle{Data and text mining}

\title[Fine-Tuning Large Neural Language Models for Biomedical Natural Language Processing]{Fine-Tuning Large Neural Language Models for Biomedical Natural Language Processing}


%
\author[R. Tinn, H. Cheng, \textit{et~al}.]{
    Robert Tinn\,$^{\text{\sfb 1,}*}$, 
    Hao Cheng\,$^{\text{\sfb 1,}*}$, 
    Yu Gu\,$^{\text{\sfb 1}}$,  
    Naoto Usuyama\,$^{\text{\sfb 1}}$,  
    Xiaodong Liu\,$^{\text{\sfb 1}}$,\\ 
    Tristan Naumann\,$^{\text{\sfb 1}}$, 
    Jianfeng Gao\,$^{\text{\sfb 1}}$ and    
    Hoifung Poon\,$^{\text{\sfb 1}**}$ 
}
\address{$^{\text{\sf 1}}$Microsoft Research, Redmond, WA  98052 USA.}

\corresp{
    $^\ast$These authors contributed equally.\\
    $^{\ast \ast}$To whom correspondence should be addressed.
}



\history{Received on XXXXX; revised on XXXXX; accepted on XXXXX}

\editor{Associate Editor: XXXXXXX}

\abstract{\textbf{Motivation:} A perennial challenge for biomedical researchers and clinical practitioners is to stay abreast with the rapid growth of publications and medical notes. Natural language processing (NLP) has emerged as a promising direction for taming information overload. In particular, large neural language models facilitate transfer learning by pretraining on unlabeled text, as exemplified by the successes of BERT models in various NLP applications.
However, fine-tuning such models for an end task remains challenging, especially with small labeled datasets, which are common in biomedical NLP. \\
\textbf{Results:} We conduct a systematic study on fine-tuning stability in biomedical NLP. We show that fine-tuning performance may be sensitive to pretraining settings, especially in low-resource domains. 
Large models have potential to attain better performance, but increasing model size also exacerbates fine-tuning instability. We thus conduct a comprehensive exploration of techniques for addressing fine-tuning instability. We show that these techniques can substantially improve fine-tuning performance for low-resource biomedical NLP applications. 
Specifically, freezing lower layers is helpful for standard BERT-$\tt BASE$ models, while layerwise decay is more effective for BERT-$\tt LARGE$ and ELECTRA models. For low-resource text similarity tasks such as BIOSSES, reinitializing the top layer is the optimal strategy. 
Overall, domain-specific vocabulary and pretraining facilitate more robust models for fine-tuning. Based on these findings, we establish new state of the art on a wide range of biomedical NLP applications.
\\
\textbf{Availability and implementation:} To facilitate progress in biomedical NLP, we release our state-of-the-art pretrained and fine-tuned models: \url{https://aka.ms/BLURB}.\\
\textbf{Contact:} hoifung@microsoft.com}



\maketitle

\section{Introduction}


Biomedical text has been growing at an explosive rate. PubMed\footnote{\url{http://www.ncbi.nlm.nih.gov/pubmed}} adds thousands of papers every day and over a million every year. 
Similarly, the digitization of patient records has created steadily growing resources of clinical text. 
For example, every year there are about two million new cancer patients in the U.S. alone, each with hundreds of clinical notes such as pathology reports and progress notes. 
By curating cutting-edge knowledge and longitudinal patient information from such text, we can potentially accelerate clinical research and improve clinical care. 
Manual curation, however, can require hours for each paper or patient, which is hard to scale given the rapid growth of biomedical text. 

Natural language processing (NLP) has emerged as a promising direction to accelerate curation by automatically extracting candidate findings for human experts to validate \citep{lu12AssistedCuration,wong21AssistedCuration}.  
However, standard supervised learning typically requires a large amount of training data. 
Consequently, task-agnostic self-supervised learning is rapidly gaining traction. By pretraining on unlabeled text, large neural language models facilitate transfer learning and have demonstrated spectacular success for a wide range of NLP applications~\citep{devlin2018bert,liu2019roberta}. 
Fine-tuning these large neural models for specific tasks, however, remains challenging, as has been shown in the general domain~\citep{griesshaber-etal-2020-fine,mosbach2021on,zhang2021revisiting}. For biomedicine, the challenge is further exacerbated by the scarcity of task-specific training data because annotation requires domain expertise and crowd-sourcing is harder to apply. For example, BIOSSES~\citep{souganciouglu2017biosses}, a semantic similarity task in the biomedical domain, contains only 100 annotated examples in total. By contrast, STS~\citep{sts-b}, a similar dataset in the general domain, contains 8,628 examples.

In this paper, we conduct a systematic study on fine-tuning stability in biomedical NLP. We ground our study in BLURB, a recently-proposed comprehensive benchmark for biomedical NLP comprising 6 tasks and 13 datasets~\citep{gu2021domainspecific}. 

We first study how pretraining settings impact fine-tuning performance. 
We show that for all applications, skipping next-sentence prediction (NSP) in pretraining has negligible effect, thus saving significant compute time, a finding consistent with general-domain observations by \citet{liu2019roberta, aroca-ouellette-rudzicz-2020-losses}. However, modeling segment IDs during pretraining may have large impact on certain semantic tasks, such as text similarity and question answering, especially when training data is scarce. Larger models (e.g., BERT-$\tt LARGE$) significantly increase fine-tuning instability, and their use often hurts downstream performance. Interestingly, changing the pretraining objective from masked language model (MLM) to ELECTRA has demonstrated improved performance in general-domain applications~\citep{clark2020electra}, but it may exacerbate fine-tuning instability in low-resource biomedical applications. 

We then conduct a comprehensive exploration of stabilization techniques to establish the best practice for biomedical fine-tuning. We show that conventional general-domain techniques, such as longer training and gradient debiasing, help but layerwise adaptation methods are key to restoring fine-tuning stability in biomedical applications. Interestingly, their efficacy may vary with pretraining settings and/or end tasks. 
For example, freezing lower layers is helpful for standard BERT-$\tt BASE$ models, whereas layerwise decay is more effective for BERT-$\tt LARGE$ and ELECTRA models. For low-resource text similarity tasks, such as BIOSSES, reinitializing the top layer is the optimal strategy. 
Overall, we find that domain-specific vocabulary and pretraining produce more robust language models. Based on these findings, we attain new state-of-the-art performance on a wide range of biomedical NLP tasks. 

Finally, we show that the best biomedical language models not only cover a much wider range of applications, but also substantially outperform  off-the-shelf biomedical NLP tools on their currently available tasks. To facilitate biomedical research and applications, we will release our state-of-the-art pretrained and task-specific fine-tuned models.




\eat{
PubMed know + RWE
Precision medicine
Manual: hours per paper, patient
NLP promising \cite{jco-cci}. No crowd-sourcing, Direct supervision
Self-supervised transfer learning, BERT model

Prior method focus on pretraining. PubMedBERT - fine-tuning, but limited
Systematic study on impact to fine-tuning efficiency/effectiveness
- Pretraining
- Vocab
- Model size
- Layer adaptation

Domain specific models achieve state of the art results on biomedical and clinical tasks \cite{gu2021domainspecific}. Recently \cite{10.1093/bioinformatics/btaa668} demonstrated how biomedical BERT models can be achieve state of the art performance on the classification of radiology text reports and event extraction \cite{10.1093/bioinformatics/btaa540}. Other approaches to mitigate the challenge of scarce data in the biomedical domain besides domain specific pre-training include multi-task learning \cite{10.1093/bioinformatics/btaa515}, however this is only possible for common tasks such as named-entity recognition.
}

\begin{methods}
\section{Methods}

In this paper, we focus our study on BERT~\citep{devlin2018bert} and its variants, which have become a mainstay of neural language models in NLP applications. In this section, we review core technical aspects in neural language model pretraining and fine-tuning, providing a basis for the key research questions of our fine-tuning study.

\subsection{Neural Language Models}
The input to a neural language model consists of text spans, such as sentences, separated by special tokens~$\tt [SEP]$. To address the problem of out-of-vocabulary words, neural language models generate a vocabulary from subword units, using Byte-Pair Encoding~\citep[BPE;][]{sennrich2015bpe} or variants such as WordPiece~\citep{kudo2018sentencepiece}.
Essentially, the BPE algorithm tries to greedily identify a small set of subwords that can compactly form all words in a given corpus. It does this by 
initializing the vocabulary with all characters and delimiters found in the corpus.  It then iteratively augments the vocabulary with a new subword that is most frequent in the corpus and can be formed by concatenating two existing subwords, until the vocabulary reaches the pre-specified size---e.g., 30,000 in standard BERT models or 50,000 in RoBERTa~\citep{liu2019roberta}. In this paper, we use the WordPiece algorithm, which is a BPE variant that augments the vocabulary using likelihood in a unigram language model rather than frequency in choosing which subwords to concatenate. 

The text corpus and vocabulary may preserve the original case ($\tt cased$) or convert all characters to lower case ($\tt uncased$). Prior work, such as \citet{gu2021domainspecific}, finds that case doesn't have significant impact on downstream tasks, so we simply use $\tt uncased$ in our work.

BERT~\citep{devlin2018bert} is a state-of-the-art neural language model based on transformer~\citep{vaswani2017attention}. 
The transformer model introduces a multi-layer, multi-head self-attention mechanism, which has demonstrated superiority in leveraging GPU computation and modeling long-range text dependencies. 
Standard BERT pretraining inputs two text spans (e.g., sentences) and assigns a distinct segment ID to each. 
The input token sequence is first processed by a lexical encoder, which combines a token embedding, a position embedding and a segment embedding by element-wise summation. 
This embedding layer is then passed to multiple layers of transformer modules. 
In each transformer layer, a contextual representation is generated for each token by summing a non-linear transformation of the representations of all tokens in the prior layer, weighted by attentions computed using a given token's representation in the prior layer as query.
The final layer outputs contextual representations for all tokens, which combines information from the whole text span.

BERT models come with two standard configurations: $\tt BASE$ uses 12 layers of transformer modules and 110 million parameters, $\tt LARGE$ uses 24 layers of transformer modules and 340 million parameters. Prior work applying BERT to biomedical NLP focuses on $\tt BASE$ models. We conduct a systematic study on $\tt LARGE$ models as well, which reveals additional challenges for fine-tuning neural language models in biomedical NLP.

\subsection{Pretraining Objectives}
Similar to other language models, the key idea of BERT pretraining is to predict held-out words in unlabeled text. Unlike most prior language models, BERT does not adhere to a generative model. Instead, \citet{devlin2018bert} introduces two self-supervised objectives: \textbf{Masked Language Model} and \textbf{Next Sentence Prediction}. 
Masked language model (MLM) randomly replaces a subset of tokens by a special token (e.g., $\tt [MASK]$), and asks the language model to predict them. 
The training objective is the cross-entropy loss between the original tokens and the predicted ones. 
Typically, 15\% of the input tokens are chosen, among which a random 80\% are replaced by $\tt [MASK]$, 10\% are left unchanged and 10\% are randomly replaced by a token from the vocabulary. 
Next sentence prediction (NSP) is a binary classification task that determines for a given sentence pair whether one sentence follows the other in the original text. 
While MLM is undoubtedly essential for BERT pretraining, the utility of NSP has been called into question in prior work~\citep{liu2019roberta}.
As such, we conduct ablation studies to probe how NSP and the use of segment IDs in pretraining might impact downstream fine-tuning performance. 

Aside from standard BERT pretraining objectives, we also consider ELECTRA~\citep{clark2020electra}, which has shown good performance in general-domain datasets such as GLUE~\citep{wang19iclr_glue} and SQuAD~\citep{rajpurkar2016squad,rajpurkar2018know}. 
ELECTRA introduces an MLM-based generator to help pretrain a discriminator for use in end tasks. Specifically, given sample masked positions, first the generator predicts the most likely original tokens as in MLM, then the discriminator classifies, for all tokens, whether each is the original one.
While ELECTRA shares some superficial similarity with generative adversarial network \citep[GAN;][]{goodfellow2014gan}, the roles of generator and discriminator are very different. After pretraining, the generator in ELECTRA is discarded and the discriminator is used for downstream fine-tuning, whereas GAN typically discards the discriminator and uses the generator. The training objective is not adversarial, but a weighted combination of MLM for the generator and classification accuracy for the discriminator. By classifying on all tokens rather than just the masked ones, ELECTRA can potentially learn more from each example while adding little overhead as the majority of compute lies in transformer layers before classification. The generator, on the other hand, does incur additional compute. Also, if the generator becomes very accurate early on, there will be little learning signal for the discriminator. Therefore, ELECTRA typically uses lower capacity in the generator compared to the discriminator (e.g., one third in $\tt BASE$ and one fourth in $\tt LARGE$ for contextual representation dimension and attention head number). 


\begin{table}[!t]
\caption{Summary of techniques for fine-tuning stabilization in recent studies and our investigation.}
\resizebox{.4791\textwidth}{!}{
\begin{tabular}{@{}ccccccc@{}}
    \hline
&  Domain & \begin{tabular}{@{}l@{}}Longer\\training\end{tabular} & \begin{tabular}{@{}l@{}}ADAM\\debiasing\end{tabular} & \begin{tabular}{@{}l@{}}Layer\\freeze\end{tabular} & \begin{tabular}{@{}l@{}}Layer-wise\\decay\end{tabular} &
\begin{tabular}{@{}l@{}}Layer\\reinit\end{tabular}\\
\midrule

\citet{griesshaber-etal-2020-fine} & \begin{tabular}{@{}l@{}}General\\(GLUE)\end{tabular} & & & \checkmark & & \\
\citet{mosbach2021on} & \begin{tabular}{@{}l@{}}General\\(GLUE)\end{tabular} & \checkmark & \checkmark & & & \\
\citet{zhang2021revisiting} & \begin{tabular}{@{}l@{}}General\\(GLUE)\end{tabular} & \checkmark & \checkmark & & \checkmark & \checkmark \\
Ours & \begin{tabular}{@{}l@{}}Biomedical\\(BLURB)\end{tabular} & \checkmark & \checkmark & \checkmark & \checkmark & \checkmark \\
\botrule
\end{tabular}}{}
\label{table:stability_summary}
\end{table}

\begin{figure}[!t]
\centering
    \includegraphics[width=0.48\textwidth]{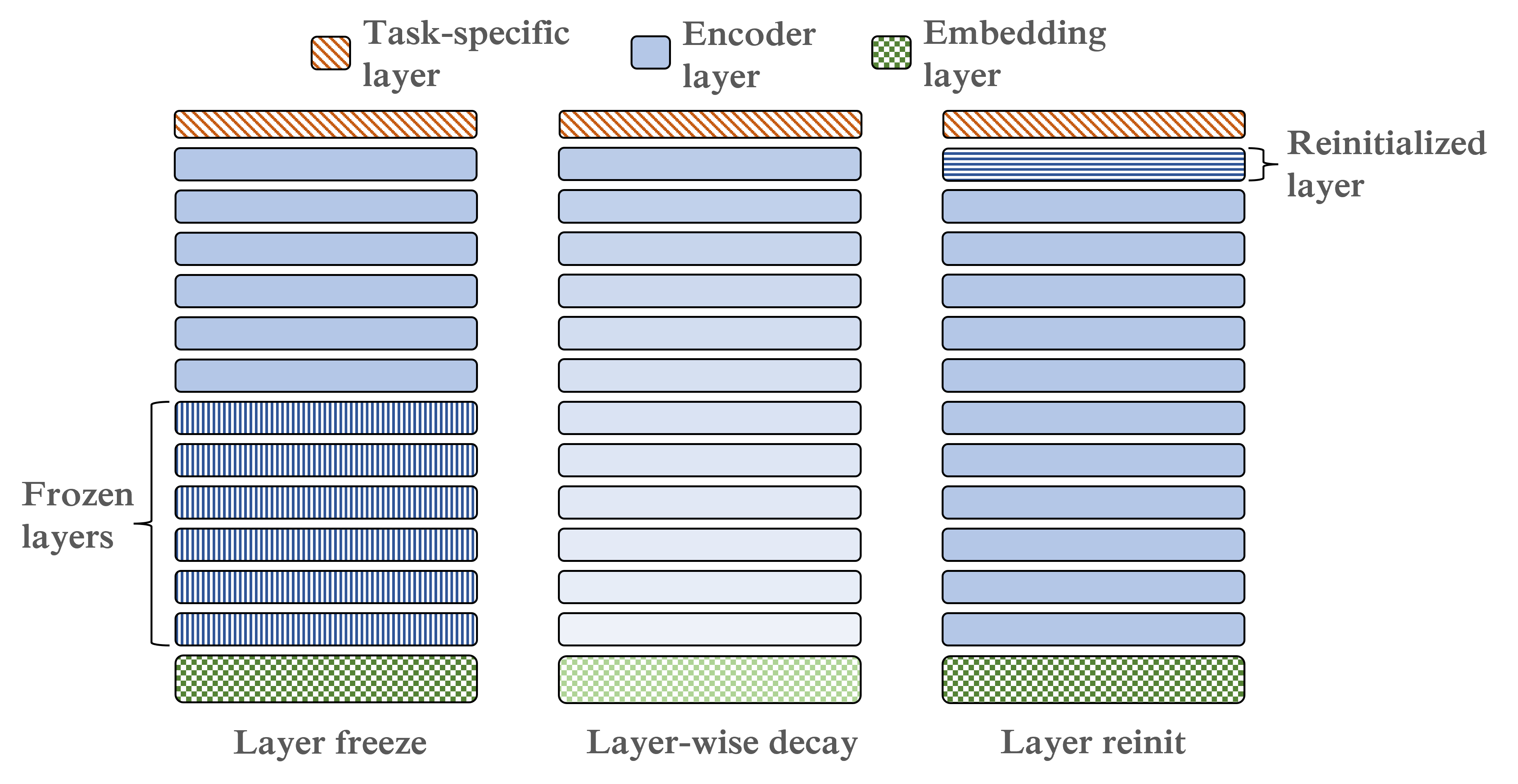}
    \caption{
    Illustration of major layer-specific adaptation methods for fine-tuning stabilization: freezing lower layers, adopting layerwise decay of learning rate, reinitializing the top layer.
    }
    \label{fig:alternate_finetune_strategies_fig}
\end{figure}

\subsection{Domain-Specific Pretraining}

The study of neural language model pretraining originates in the general domain, including newswire and web. For example, the original BERT model was pretrained on Wikipedia and BooksCorpus~\citep{devlin2018bert}. RoBERTa~\citep{liu2019roberta}, another representative BERT model, was pretrained on a larger web corpus. 
Biomedical text is quite different from general domain text and domain-specific pretraining has been shown to substantially improve performance in biomedical NLP applications~\citep{lee2020biobert,peng2019transfer,gu2021domainspecific}. In particular, \citet{gu2021domainspecific} conducted a thorough analysis on domain-specific pretraining, which highlights the utility of using a domain-specific vocabulary and pretraining on domain-specific text from scratch. 
We build on their work and study how domain-specific pretraining might impact fine-tuning stability, especially for larger models and/or with alternative pretraining settings. To facilitate our investigation, we pretrained PubMedBERT-$\tt LARGE$ and PubMedELECTRA ($\tt BASE$ and $\tt LARGE$) following the same setting of PubMedBERT ($\tt BASE$) in \citet{gu2021domainspecific}.

\eat{
\hoifung{
moving Fig/Tbl to top of column?

Fig 1: text and line look a bit faint; task-specific layer color didn't seem to match; "Freezing layer" -> "Freezing-layer"; "Removing layer" -> "Removing-layer"

Remove "Encoder layer" from legend: this suggests that other layers are not encoder, but they are. 

"Embedding layer" -> Embedding
"Task-specific layer" -> "Task-specific prediction"
}
}


\subsection{Fine-Tuning Stability}



Prior work studying fine-tuning stability and mitigation methods tends to focus on general domain---e.g., using BERT models pretrained on general-domain corpora and evaluating on GLUE~\citep{wang19iclr_glue} or SuperGLUE~\citep{superglue2019_8589}. \autoref{table:stability_summary} summarizes representative recent work and common stabilization techniques. 
Small adjustments to the conventional optimization process may have surprisingly significant effect. For example,  \citet{mosbach2021on} and \citet{zhang2021revisiting} show that simply training for longer time helps reduce fine-tuning instability with small training datasets. They also show that bias correction, which was proposed in the original ADAM algorithm~\citep{kingma2014adam} but was not used in fine-tuning from the original BERT paper~\citep{devlin2018bert}, can enhance fine-tuning stability by effectively reducing learning rates in the first few iterations. 

Such minor adaptations are already highly effective for general-domain applications~\citep{mosbach2021on}. However, biomedical datasets are often much smaller than their general-domain counterparts. For example, as aforementioned for text similarity, the biomedical dataset BIOSSES~\citep{souganciouglu2017biosses} is much smaller than the general-domain dataset STS~\citep{sts-b}. Similarly, the question-answering datasets in BLURB have only a few hundred instances, compared to over one hundred thousand in SQuAD~\citep{rajpurkar2016squad}. 

Therefore, we systematically study advanced layer-specific adaptation techniques previously studied in the general domains: freezing pretrained parameters in the lower layers~\citep{griesshaber-etal-2020-fine}, adopting layerwise learning-rate decay \citep{clark2020electra}, and reinitializing parameters in the top layer \citep{zhang2021revisiting}. 
See \autoref{fig:alternate_finetune_strategies_fig}.
Essentially, these techniques represent various ways to alleviate the vanishing gradient problem in training deep neural networks~\citep{singh2015layer}, 
where optimization suffers from severe ill conditioning and requires adapting learning rates for individual layers. 
Interestingly, we find that their efficacy may interact with the pretraining setting and the end task.

\eat{
Standard task-specific fine-tuning will take the top BERT layer as the meaning representation and pass it to the task-specific prediction component (e.g., a linear classification layer). 

review strategies - show graphs

It has been shown that BERT-based transformer models can be down-scaled whilst maintaining strong performance on NLP tasks \citep{sajjad2020poor}. Down-scaling these models has multiple advantages such as reducing fine-tuning times and memory usage. Figure~1\vphantom{\ref{fig:alternate_finetune_strategies_fig}} illustrates two strategies we apply to improve the efficiency of fine-tuning. Biomedical NLP practitioners often have to contend with computationally limited environments, particularly if a data set must be contained within a secure environment to mitigate privacy issues.

\hoifung{We could argue about down-scaling for efficiency; but think we should first highlight the more important bit of stability. 

What are the the best citations for stability? 

What about training longer and layer reinit? Cf. Fig 3.

Here we talked about removing. Did we try that in Fig 3?
}
}

\end{methods}


\section{Results}




In this section, we conduct a systematic study on fine-tuning stability and mitigation methods in the presence of various pretraining settings and large models. 
We ground our study on the Biomedical Language Understanding \& Reasoning Benchmark \citep[BLURB;][]{gu2021domainspecific}. BLURB is a comprehensive benchmark for biomedical NLP, spanning six tasks and thirteen datasets, including applications with very small training datasets, such as text similarity and question answering. 
To facilitate a head-to-head comparison, we follow the train/dev/test setup from BLURB in all our experiments.

\subsection{Instability with Alternative Pretraining Settings}


We first conduct an ablation study to evaluate the impact of pretraining settings on fine-tuning stability. Prior work on fine-tuning stability focuses almost exclusively on $\tt LARGE$ models; we show that $\tt BASE$ models also suffer instability if we deviate from standard BERT pretraining settings.

Specifically, we experiment with skipping next-sentence prediction (NSP) during pretraining. 
Standard BERT pretraining inputs two text sequences (with two distinct segment IDs). 
We also experiment with inputting a single sequence at a time (with same segment ID). 
For a head-to-head comparison, we pretrained all language models from scratch on PubMed abstracts (i.e., using the same settings as PubMedBERT) and adopted the same fine-tuning settings as in \citet{gu2021domainspecific}. 

\autoref{tab:pretraining} shows the results. In general, NSP has relatively little impact on end task performance. However, pretraining with single sequences leads to a substantial performance drop in the sentence similarity task (BIOSSES). Presumably performance degrades because this task requires comparison of two sentences and the training set is very small, therefore pretraining with two text segments helps. Surprisingly, pretraining with single sequences substantially improves test performance on PubMedQA, even though the task also inputs two text segments. Interestingly, even with the original pretraining setting (with NSP and two segments), simply using a single segment ID in fine-tuning for PubMedQA would result in a similarly large gain in test performance (F1 63.92, not shown in the table). However, the standard setting (using separate segment ID) is still better for BIOSSES and BioASQ. 


We also evaluate using the ELECTRA objective. Unlike in the general domain, ELECTRA does not show clear improvements over the MLM objective in the biomedical domain. In fact, ELECTRA performs worse on most tasks and suffers a catastrophic performance drop in text similarity. We note that these tasks also happen to have relatively small training sets. This may appear contradictory with some recent work that demonstrates superior results using ELECTRA in biomedical NLP~\citep{kanakarajan-etal-2021-bioelectra}. Later, we show that it is indeed possible to attain higher performance with ELECTRA, but doing so requires various techniques to stabilize and improve fine-tuning. Compared to BERT with the MLM objective, ELECTRA is generally more difficult to fine-tune and demonstrates no significant advantage for biomedical NLP.

\begin{table}[!t]
\processtable{
Comparison of BLURB test performance with various pretraining settings: standard BERT pretraining; BERT pretraining without NSP (i.e., MLM only); BERT pretraining with MLM only and single-sequence (single segment ID); ELECTRA.
\label{tab:pretraining}
} 
{
\begin{tabular}{@{}lccccc@{}}
\midrule
&  BERT & \begin{tabular}{@{}l@{}}BERT\\(no NSP)\end{tabular} & \begin{tabular}{@{}l@{}}BERT (no NSP,\\single seq)\end{tabular} & ELECTRA \\
\midrule 
BC5-chem     &     \textbf{93.33} &     93.21 &            93.20 &    93.00 \\
BC5-disease  &     \textbf{85.62} &     85.29 &            85.44 &    84.84 \\
NCBI-disease &     87.82 &     88.29 &            \textbf{88.68} &    87.17 \\
BC2GM        &     84.52 &     84.41 &            \textbf{84.63} &    84.03 \\
JNLPBA       &     \textbf{79.10} &     79.01 &            \textbf{79.10} &    78.57 \\
\midrule 
EBM PICO     &     73.38 &     \textbf{73.87} &            73.64 &    73.57 \\
\midrule 
ChemProt     &     \textbf{77.24} &     76.82 &            76.88 &    76.34 \\
DDI          &     82.36 &     \textbf{82.64} &            82.45 &    80.58 \\
GAD          &     \textbf{83.96} &     82.30 &            83.24 &    83.40 \\
\midrule 
BIOSSES      &     \textbf{93.46} &     93.12 &            75.50 &    80.24 \\
\midrule 
HoC          &     82.32 &     \textbf{82.37} &            81.91 &    81.28 \\
\midrule 
PubMedQA     &     55.84 &     56.40 &            \textbf{66.66} &    64.96 \\
BioASQ       &     87.56 &     83.57 &            85.64 &    \textbf{88.93} \\
\midrule 
\begin{tabular}{@{}l@{}}BLURB\\score\end{tabular}   &     \textbf{81.35} &     81.00 &            79.04 &    79.61 \\
\botrule
\end{tabular}
}{}
\end{table}


\begin{table}[t]
\processtable{
Ablation study on optimization adjustments in fine-tuning by comparing BIOSSES test performance under various pretraining settings. Improved optimization used bias correction in ADAM and up to 100 epochs in fine-tuning (vs. up to five epochs in standard setting), all with $\tt BASE$ models.
\label{tab:bias}
} 
{\begin{tabular}{l|ccc}
\toprule
Pretraining Setting 
& \begin{tabular}{@{}l@{}}Improved\\optimization\end{tabular} 
& \begin{tabular}{@{}l@{}}Standard  \\epochs\end{tabular} 
& \begin{tabular}{@{}l@{}}No bias\\correction\end{tabular} \\
\midrule
BERT &  \textbf{93.46} &      92.64 &    91.75 \\ 
BERT (no NSP) &  \textbf{93.12} &      91.31 &    92.35 \\ 
BERT (no NSP, single seq) &   \textbf{75.50} &       0.65 &     70.50 \\ 
ELECTRA &  80.24 &      49.87 &    \textbf{80.41} \\ 
\bottomrule
\end{tabular}}{}
\end{table}

\subsection{Stabilization by Adjusting Standard Optimization}


As previously mentioned, prior studies conclude that small optimization adjustments often suffice to restore fine-tuning stability in $\tt LARGE$ models. In biomedical NLP, however, we found that such adjustments are necessary to prevent catastrophic performance drops, but are not always sufficient for stabilizing fine-tuning, even with $\tt BASE$ models. 
\autoref{tab:bias} shows an ablation study on BIOSSES. 
In this case, forgoing either adjustment leads to a significant performance drop. 
But, as noted in the last subsection, even if both are used, fine-tuning remains unstable with alternative pretraining settings, which requires more advanced stabilization techniques.


\begin{table}[!t]
\processtable{
Comparison of test performance on Sentence Similarity (SS) and Question Answering (QA) tasks with major layer-specific adaptation methods, all with $\tt BASE$ models.
\label{tab:base_stability}
} 
{\begin{tabular}{l|cc|cc|cc|cc}
\toprule
\begin{tabular}{@{}l@{}}Pretraining\\setting\end{tabular} & \multicolumn{2}{c}{Baseline} & \multicolumn{2}{c}{\begin{tabular}{@{}l@{}}Layer\\freeze\end{tabular}} & \multicolumn{2}{c}{\begin{tabular}{@{}l@{}}Layerwise\\decay\end{tabular}} & \multicolumn{2}{c}{\begin{tabular}{@{}l@{}}Layer\\reinit\end{tabular}} \\
\midrule
&       SS &         QA &           SS &         QA &               SS &         QA &           SS &         QA \\
\midrule
BERT &    93.5 &       71.7 &        92.9 &      \textbf{76.3} &            93.4 &     74.3 &        \textbf{94.5} &      69.3 \\
BERT (no NSP) &    93.1 &     70.0 &        \textbf{94.0} &      \textbf{73.2} &            93.0 &     71.8 &        92.9 &      69.3 \\
\begin{tabular}{@{}l@{}}BERT (no NSP,\\single seq)\end{tabular} &     75.5 &      76.2 &        72.1 &      \textbf{78.0} &            74.1 &      77.5 &        \textbf{85.0} &     72.8 \\
ELECTRA &    80.2 &     77.0 &        83.1 &      78.7 &            83.6 &      \textbf{79.1} &        \textbf{88.7} &      74.7 \\
\bottomrule
\end{tabular}}{}
\processtable{
Comparison of test performance on Sentence Similarity (SS) and Question Answering (QA) tasks with layer-specific adaptation methods on $\tt LARGE$ models.
\label{tab:large_stability}
} 
{
\begin{tabular}{l|cc|cc|cc|cc}
\toprule
 & \multicolumn{2}{c}{Baseline} & \multicolumn{2}{c}{\begin{tabular}{@{}l@{}}Layer\\freeze\end{tabular}} & \multicolumn{2}{c}{\begin{tabular}{@{}l@{}}Layerwise\\decay\end{tabular}} & \multicolumn{2}{c}{\begin{tabular}{@{}l@{}}Layer\\reinit\end{tabular}} \\
\midrule
&       SS &         QA &           SS &         QA &               SS &         QA &           SS &         QA \\
\midrule
BioBERT-$\tt LARGE$ &     84.9 &       60.9 &         88.7 &       65.4 &             90.1 &       \textbf{67.0} &         \textbf{91.5} &       64.8 \\       
BlueBERT-$\tt LARGE$ &     82.4 &       59.9 &         84.6 &       62.7 &             84.8 &       \textbf{63.7} &         \textbf{86.2} &       63.0 \\       
\begin{tabular}{@{}l@{}}PubMedBERT\\-$\tt LARGE$\end{tabular} &     91.1 &       77.8 &         91.2 &       79.3 &             90.9 &       \textbf{80.4} &         \textbf{92.7} &       76.9 \\
\begin{tabular}{@{}l@{}}PubMedELECTRA\\-$\tt LARGE$\end{tabular} &     71.6 &       67.6 &         86.4 &       76.5 &             86.2 &       \textbf{79.1} &         \textbf{90.3} &       76.0 \\
\bottomrule
\end{tabular}
}{}
\end{table}

\subsection{Stabilization by Layer-Specific Adaptation}

Next, we study various layer-specific adaptation methods in fine-tuning. 
Given that most models suffer from high instability on sentence similarity (BIOSSES) and question answering (BioASQ and PubMedQA), we focus on those tasks. For question answering, we report the mean performance.
\autoref{tab:base_stability} shows the results. 
All three methods are broadly beneficial, but their effects vary substantially with tasks and pretraining settings. Freezing lower layers is helpful for BERT models with the standard MLM objective, whereas layerwise decay is more effective for ELECTRA models. For sentence similarity, reinitializing the top layer is the optimal strategy. We focus our study on sentence similarity and question answering tasks, as other datasets in BLURB are relatively large and do not suffer from stability issues. We explored a combination of layer-specific adaptation methods but found little gain in preliminary experiments.

\subsection{Stabilization for Larger Models}

It is well known that larger models are more finicky to fine-tune \citep{devlin2018bert}. 
Again, we focus on sentence similarity (BIOSSES) and question answering (BioASQ and PubMedQA). 
Indeed, we observe a substantial drop in test performance on sentence similarity and question answering tasks for most large models (see \autoref{tab:large_stability}). 
Note that to avoid clutter, we only show the average scores for the question-answering tasks.

Surprisingly, PubMedBERT-$\tt LARGE$ is a notable exception because it doesn't suffer any catastrophic performance drop. In fact, it actually gains slightly on the question-answering tasks. This stands in stark contrast with other models such as BioBERT~\citep{lee2020biobert} and BlueBERT~\citep{peng2019transfer}. We hypothesize that its robustness stems from domain-specific vocabulary and pretraining. Interestingly, although PubMedELECTRA-$\tt LARGE$ is also pretrained in the same domain-specific fashion, it suffers a similar performance drop, which provides further evidence that the ELECTRA pretraining objective may exacerbate fine-tuning instability. 

Optimization adjustments (longer training time and ADAM bias correction) have been used in all these experiments. Unlike in the general domain, they are not sufficient to restore stability. As in the case of $\tt BASE$ models, layer-specific adaptation methods can substantially reduce fine-tuning instability, in some cases enabling $\tt LARGE$ models to attain even higher performance than $\tt BASE$ (e.g., PubMedBERT-$\tt LARGE$ on QA and PubMedELECTRA-$\tt LARGE$ on SS). See \autoref{tab:large_stability}. 

Like \citet{gu2021domainspecific}, we also observe that domain-specific vocabulary and pretraining are far superior,  
as PubMedBERT-$\tt LARGE$ substantially outperforms BioBERT-$\tt LARGE$ \citep{lee2020biobert} and BlueBERT-$\tt LARGE$ \citep{peng2019transfer}.
Again, while ELECTRA models can perform reasonably well with advanced stabilization techniques, they are still finicky to fine-tune and are not superior over BERT models with the standard MLM pretraining objective. As with $\tt BASE$ models, reinitializing the top layer is still the optimal strategy for sentence similarity. However, for question answering, layerwise decay is superior for $\tt LARGE$ models.

\autoref{tab:large_models_blurb} compares overall BLURB test performance for $\tt LARGE$ models with both improved optimization and layer-specific adaptation. 
They help stabilize fine-tuning, with no $\tt LARGE$ model suffers significant instability issues. 
With domain-specific vocabulary and pretraining, PubMedBERT-$\tt LARGE$ and PubMedElectra-$\tt LARGE$ benefit the most and attain significant gain over $\tt BASE$.

\begin{table}[!t]
\processtable{
Comparison of BLURB test performance using $\tt LARGE$ models (24 layers, 300M+ parameters), with optimal layer-specific stabilization methods.
\label{tab:large_models_blurb}
} 
{\begin{tabular}{lcccc}
\toprule
& \multirow{2}{*}{\begin{tabular}{@{}l@{}}BioBERT\\-$\tt LARGE$\end{tabular}} &  \multirow{2}{*}{\begin{tabular}{@{}l@{}}BlueBERT\\-$\tt LARGE$\end{tabular}} &  \multirow{2}{*}{\begin{tabular}{@{}l@{}}PubMedBERT\\-$\tt LARGE$\end{tabular}} &  \multirow{2}{*}{\begin{tabular}{@{}l@{}}PubMedELECTRA\\-$\tt LARGE$\end{tabular}} \\
& & & & \\
\midrule
BC5-chem     &          93.05 &           90.24 &             \textbf{93.23} &                92.90 \\
BC5-disease  &          84.97 &           82.93 &             \textbf{85.77} &                84.82 \\
NCBI-disease &          \textbf{88.76} &           86.44 &             88.25 &                87.93 \\
BC2GM        &          84.21 &           80.86 &             \textbf{84.72} &                83.87 \\
JNLPBA       &          78.83 &           77.59 &             \textbf{79.44} &                78.77 \\
\midrule
EBM PICO     &          73.81 &           72.43 &             73.61 &                \textbf{73.95} \\
\midrule
ChemProt     &          77.79 &           71.31 &             \textbf{78.77} &                76.80 \\
DDI          &          81.53 &           78.99 &             \textbf{82.39} &                78.92 \\
GAD          &          82.47 &           75.80 &             83.57 &                \textbf{83.93} \\
\midrule
BIOSSES      &          91.53 &           86.18 &             \textbf{92.73} &                90.33 \\
\midrule
HoC          &          81.57 &           81.35 &             \textbf{82.57} &                82.37 \\
\midrule
PubMedQA     &          55.16 &           55.24 &             \textbf{67.38} &                65.02 \\
BioASQ       &          78.93 &           72.21 &             \textbf{93.36} &                93.14 \\
\midrule
BLURB score  &          80.09 &           77.11 &             \textbf{82.86} &                81.88 \\
\midrule
\begin{tabular}{@{}l@{}} $\Delta$ $\tt BASE$\\model\end{tabular}         &          -0.59 &            +0.15 &              +0.58 &                 +0.37 \\
\bottomrule
\end{tabular}}{}
\end{table}


\begin{table}[!t]
\processtable{
Ablation study on the impact of model pruning by comparing test performance on BLURB tasks after removing top layers of PubMedBERT.
\label{tab:removed_layers}
} 
{
\begin{tabular}{lcccc|c}
\toprule
 & \multicolumn{4}{c}{Layers removed } & Performance \\
& 0 & 2 & 4 & 6 & Drop \\
\midrule
BC5-chem        & 93.3 & 93.2 & 93.0 & 92.4 &  -0.93 \\
BC5-disease     & 85.6 & 85.4 & 85.2 & 84.3 &  -1.33 \\
NCBI-disease    & 87.8 & 88.4 & 88.0 & 87.4 &  -0.44 \\
BC2GM           & 84.5 & 84.3 & 83.5 & 82.1 &  -2.47 \\
JNLPBA          & 79.1 & 78.9 & 78.9 & 78.1 &  -1.03 \\
\midrule
EBM PICO        & 73.4 & 73.4 & 73.3 & 73.4 &  -0.05 \\
\midrule
ChemProt        & 77.2 & 76.1 & 73.4 & 72.7 &  -4.50 \\
DDI             & 82.4 & 82.2 & 79.7 & 79.3 &  -3.06 \\
GAD             & 84.0 & 82.3 & 80.2 & 79.2 &  -4.75 \\
\midrule
BIOSSES         & 92.3 & 92.7 & 92.8 & 92.1 &  -0.18 \\
\midrule
HoC             & 82.3 & 82.5 & 82.4 & 82.0 &  -0.31 \\
\midrule
PubMedQA        & 55.8 & 51.2 & 49.8 & 50.1 &  -6.08 \\
BioASQ          & 87.6 & 83.7 & 77.5 & 74.0 & -13.56 \\
\bottomrule
\end{tabular}}{}
\end{table}

\subsection{Ablation Study on Layer Removal}

Rising concerns about computation cost of large pretrained models have spawned research in model pruning, such as removing top layers of a BERT model~\citep{sajjad2020poor}. We thus conducted an ablation study on BLURB tasks to assess the impact of removing top layers from PubMedBERT. 
\autoref{tab:removed_layers} shows the results. Indeed, pruning barely impacts fine-tuning efficacy for many tasks, such as named entity recognition, evidence-based medical information extraction, sentence similarity, and document classification. Test performance does not substantially drop even when the top half of the layers were removed, suggesting that these tasks are relatively easy and do not require deep semantic modeling. By contrast, test performance in relation extraction and question answering was substantially impacted by layer removal, dropping up to 3-4 absolute points for the former and up to 6-13 points for the latter. This suggests model pruning may make sense for simpler tasks, but not for semantically more challenging tasks. 

\subsection{New State of the Art in Biomedical NLP}


To further improve test performance for low-resource tasks, a common technique is to combine the training set and development set to train the final model---after hyperparameter search is done. We found that for most biomedical NLP tasks, this was not necessary, but it had a significant effect on BIOSSES. This is not surprising given that this dataset is the smallest.


By combining all our findings on optimal fine-tuning strategy, we establish a new state of the art in biomedical NLP. 
\autoref{tab:strategic_blurb} and \autoref{tab:strategic_blurb_large} show the results. PubMedBERT with the MLM pretraining objective remains the best model, consistently outperforming ELECTRA in most tasks, although the latter does demonstrate some advantage in question answering tasks, as can be seen in its superior performance with $\tt BASE$ models. With more extensive hyperparameter tuning, the gap between $\tt BASE$ and $\tt LARGE$ is smaller, compared to more standard fine-tuning 
(\autoref{tab:large_stability}), which is not surprising. Overall, we were able to significantly improve the BLURB score by 1.6 absolute points, compared to the original PubMedBERT results in \citet{gu2021domainspecific} (from 81.35 to 82.91).


\subsection{Comparison with Off-the-Shelf Tools}

While there are many off-the-shelf tools for general-domain NLP tasks, there are few available for the biomedical domain. An exception is scispaCy~\citep{Neumann2019ScispaCyFA}, with a limited scope focusing on named entity recognition (NER).
\autoref{table:scispacy_original} compares sciSpaCy performance with PubMedBERT on BLURB NER tasks. scispaCy comes with two versions, trained on JNLPBA and BC5CDR, 
respectively. 
We compare individually and to an oracle version that picks the optimal between the two for each evaluation dataset. While scispaCy performs well, PubMedBERT fine-tuned models attain substantially higher scores. We note that many scispaCy errors stem from imperfect entity boundaries. We thus further compare the two using a lenient score that regards overlapping predictions as correct (\autoref{table:scispacy}).
As expected, the gap shrinks but PubMedBERT models still demonstrate overwhelming improvement, raising the average score by over 10 points.

\begin{table}[!t]
\processtable{Comparison of BLURB test performance with $\tt BASE$ models: standard fine-tuning vs optimal fine-tuning with advanced stabilization methods PLUS extensive hyperparameter search.
\label{tab:strategic_blurb}
}{
\begin{tabular}{l|cc|cc}
\toprule
& \multicolumn{2}{c}{PubMedBERT} & \multicolumn{2}{c}{PubMedELECTRA}\\
\toprule
Fine-tuning & Standard & Optimal & Standard & Optimal\\
\midrule
BC5-chem     &       \textbf{93.33} &       \textbf{93.33} &          93.19 &          93.32 \\
BC5-disease  &       \textbf{85.62} &       \textbf{85.62} &          84.99 &          85.16 \\
NCBI-disease &       87.82 &       \textbf{88.21} &          87.68 &          87.73 \\
BC2GM        &       84.52 &       \textbf{84.55} &          83.79 &          83.79 \\
JNLPBA       &       79.10 &       \textbf{79.16} &          78.60 &          78.64 \\
\midrule
EBM PICO     &       73.38 &       73.45 &          73.70 &          \textbf{73.72} \\
\midrule
ChemProt     &       77.24 &       \textbf{77.41} &          76.54 &          76.74 \\
DDI          &       82.36 &       \textbf{83.17} &          80.58 &          81.09 \\
GAD          &       83.96 &       \textbf{84.01} &          83.40 &          83.98 \\
\midrule
BIOSSES      &       92.30 &       \textbf{94.49} &          80.24 &          92.01 \\
\midrule
HoC          &       82.32 &       \textbf{83.02} &          81.45 &          82.57 \\
\midrule
PubMedQA     &       55.84 &       63.92 &          \textbf{67.64} &          \textbf{67.64} \\
BioASQ       &       87.56 &       91.79 &          87.71 &          \textbf{92.07} \\
\midrule
\begin{tabular}{@{}l@{}}BLURB\\score\end{tabular}  &       81.16 &       \textbf{82.75} &          79.81 &          82.41 \\
\botrule
\end{tabular}}{}
\processtable{Comparison of BLURB test performance with $\tt LARGE$ models: standard fine-tuning vs optimal fine-tuning with advanced stabilization methods PLUS extensive hyperparameter search.
\label{tab:strategic_blurb_large}
} {
\begin{tabular}{l|cc|cc}
\toprule
& \multicolumn{2}{c}{PubMedBERT} & \multicolumn{2}{c}{PubMedELECTRA}\\
\toprule
Fine-tuning & Standard & Optimal & Standard & Optimal\\
\midrule
BC5-chem     &             \textbf{93.23} &             \textbf{93.23} &                92.90 &                93.25 \\
BC5-disease  &             \textbf{85.77} &             \textbf{85.77} &                84.82 &                85.23 \\
NCBI-disease &             \textbf{88.25} &             \textbf{88.25} &                87.93 &                88.19 \\
BC2GM        &             \textbf{84.72} &             \textbf{84.72} &                83.87 &                84.47 \\
JNLPBA       &             \textbf{79.44} &             \textbf{79.44} &                78.77 &                78.77 \\
\midrule
EBM PICO     &             73.61 &             73.61 &                73.95 &                \textbf{74.02} \\
\midrule
ChemProt     &             \textbf{78.77} &             \textbf{78.77} &                76.80 &                77.26 \\
DDI          &             82.39 &             \textbf{82.78} &                78.92 &                80.37 \\
GAD          &             83.57 &             83.76 &                \textbf{83.93} &                \textbf{83.93} \\
\midrule
BIOSSES      &             90.29 &             \textbf{92.73} &                86.17 &                92.69 \\
\midrule
HoC          &             82.57 &             \textbf{82.70} &                82.37 &                82.37 \\
\midrule
PubMedQA     &             63.18 &             \textbf{67.38} &                60.18 &                65.02 \\
BioASQ       &             92.36 &             \textbf{93.36} &                81.71 &                93.14 \\
\midrule
\begin{tabular}{@{}l@{}}BLURB\\score\end{tabular}  &             82.02 &             \textbf{82.91} &                79.83 &                82.44 \\
\botrule
\end{tabular}}{}
\end{table}

\begin{table}[!t]
\processtable{Comparison of PubMedBERT fine-tuned models and scispaCy on BLURB named entity recognition tasks (standard entity-level test F1 score).
\label{table:scispacy_original}
}{\begin{tabular}{@{}lccc|cc@{}}
\toprule
& \multicolumn{3}{c}{scispaCy} & \multicolumn{2}{c}{PubMedBERT}\\
& jnlpba & bc5cdr & max & $\tt BASE$ & $\tt LARGE$\\
\midrule
BC5-chem          & 3.60 &  86.49 &     86.49 & \textbf{93.33} &     93.23 \\      
BC5-disease       &   1.35 & 80.03 &     80.03 & 85.62 &     \textbf{85.77} \\      
NCBI-disease      &   1.77 &   57.18 &     57.18 & 87.82 &     \textbf{88.25} \\      
BC2GM             &  51.98 &      6.73 &     51.98 & 84.52 &     \textbf{84.72} \\  
JNLPBA            &   77.31 &     10.28 &     77.31 & 79.10 &     \textbf{79.44} \\
\midrule
Mean Score & 27.20 &    48.14 &     70.60 & 86.08 &     \textbf{86.28} \\
\botrule
\end{tabular}}{}
\end{table}


\begin{table}[!t]
\processtable{Comparison of PubMedBERT fine-tuned models and scispaCy on BLURB named entity recognition tasks (relaxed entity-level test F1 score - overlap counted as correct).
\label{table:scispacy}
} {\begin{tabular}{@{}lccc|cc@{}}
\toprule
& \multicolumn{3}{c}{scispaCy} & \multicolumn{2}{c}{PubMedBERT}\\
& jnlpba & bc5cdr & max & $\tt BASE$ & $\tt LARGE$\\
\midrule
BC5-chem          & 7.70 &  91.42 &     91.42 & 95.18 &     \textbf{95.37} \\      
BC5-disease       &   2.09 &88.88 &     88.88 & 93.34 &     \textbf{93.74} \\      
NCBI-disease      &   12.94 &   74.64 &     74.64 & 95.22 &     \textbf{95.24} \\      
BC2GM             &  68.87 &      15.92 &     68.87 & 95.56 &     \textbf{96.05} \\  
JNLPBA            &   87.25 &     20.50 &     87.25 & 88.79 &     \textbf{88.81} \\
\midrule
Mean Score & 35.77 &    58.27 &     82.21 & 93.62 &     \textbf{93.84} \\
\botrule
\end{tabular}}{}
\end{table}

\section{Discussion}

Studies on pretraining and fine-tuning large neural language models originated in general domain, such as newswire and the web. 
Recently, there has been increasing interest in biomedical pretraining \citep{lee2020biobert,peng2019transfer,alsentzer-etal-2019-publicly,gu2021domainspecific} and applications \citep{10.1093/bioinformatics/btaa668,10.1093/bioinformatics/btaa540}. In particular, \citet{gu2021domainspecific} conducted an extensive evaluation of pretrained models on wide-ranging biomedical NLP tasks. However, they focus on domain-specific pretraining, whereas we study fine-tuning techniques and explore how they might interact with tasks and pretraining settings.

Prior studies on fine-tuning stability focus on general domain and $\tt LARGE$ models, and often conclude that simple optimization adjustments, such as longer training time and ADAM debiasing, suffice for stabilization. By contrast, we show that in biomedical NLP, even $\tt BASE$ models may exhibit serious instability issues and simple optimization adjustments are necessary, but not sufficient, to restore stabilization. We systematically study how fine-tuning instability may be exacerbated with alternative pretraining settings such as using single sequences and the ELECTRA objective. We show that layer-specific adaptation methods help substantially in stabilization and identify the optimal strategy based on tasks and pretraining settings. 

Multi-task learning can also mitigate the challenge presented by low-resource biomedical tasks~\citep{10.1093/bioinformatics/btaa515}, but doing so generally requires applications with multiple related datasets, such as named-entity recognition (NER). As we can see from prior sections, NER tasks are relatively easy and domain-specific pretrained models can attain high scores without specific adaptation. 

Other relevant methods include Mixout~\citep{Lee2020Mixout:}, which hasn't been found to consistently improve performance, and fine-tuning on intermediate tasks~\citep{pruksachatkun-etal-2020-intermediate}, which is not always applicable and incurs substantial computation. We focus on layer-specific adaptation techniques that are generalizable and easily implemented. 

Adversarial training can also help instability issues \citep{jiang-etal-2020-smart,zhu2020freelb,cheng-etal-2021-posterior} and prompt-based learning has been shown to work well in low-resource settings \citep{schick-schutze-2021-just,gao-etal-2021-making}. We leave the exploration of these techniques in biomedical NLP to future work.

\section{Conclusion}

We present a comprehensive study on fine-tuning large neural language models for biomedical NLP applications. We show that fine-tuning instability is prevalent for low-resource biomedical tasks and is further exacerbated with alternative pretraining settings and $\tt LARGE$ models. In thorough evaluation of optimization adjustments and layer-specific adaptation techniques, we identify the best practice for fine-tuning stabilization, establishing new state of the art in the BLURB benchmark for biomedical NLP. Future directions include: applications to other biomedical tasks; exploring fine-tuning stability in clinical NLP; further study on model pruning and compression for practical use cases.

\section*{Data Availability}
The BLURB benchmark underlying this article is available at \url{https://aka.ms/BLURB}.
The models underlying this article will be made available at \url{https://huggingface.co/microsoft}.



\vspace{12pt}
\noindent\textit{Conflict of Interest}: none declared.


\bibliographystyle{natbib}
\bibliography{ref_rob,ref_tristan,ref_datasets}
\end{document}